\documentclass[11pt]{article}
\usepackage{graphicx}
\usepackage{geometry}
\usepackage{cite}
\usepackage{subcaption}
\usepackage[export]{adjustbox}
\usepackage{amsmath, amssymb, amsfonts}
\usepackage{textcomp}
\usepackage{xcolor}
\usepackage{hyperref}
\def\BibTeX{{\rm B\kern-.05em{\sc i\kern-.025em b}\kern-.08em
    T\kern-.1667em\lower.7ex\hbox{E}\kern-.125emX}}
\title{
A Low-Cost UAV Deep Learning Pipeline for Integrated Apple Disease Diagnosis,\\
Freshness Assessment, and Fruit Detection
}

\author{
Soham Dutta$^{1}$,
Soham Banerjee$^{1}$,
Sneha Mahata$^{2}$,
Anindya Sen$^{1}$,
Sayantani Datta$^{1}$ \\[0.5em]
$^{1}$Department of Electronics and Communication Engineering,\\
Heritage Institute of Technology, Kolkata, West Bengal, India \\[0.3em]
$^{2}$Department of Computer Science and Engineering (AI \& ML),\\
Heritage Institute of Technology, Kolkata, West Bengal, India \\[0.3em]
}

\date{}

\begin{document}
\maketitle
\begin{abstract}
Apple orchards require timely disease detection, fruit quality assessment, and yield estimation, yet existing UAV-based systems address such tasks in isolation and often rely on costly multispectral sensors. This paper presents a unified, low-cost RGB-only UAV-based orchard intelligent pipeline integrating ResNet-50 for leaf disease detection, VGG-16 for apple freshness determination, and YOLOv8 for real-time apple detection and localization. The system runs on an ESP32-CAM and Raspberry Pi, providing fully offline on-site inference without cloud support. Experiments demonstrate 98.9\% accuracy for leaf disease classification, 97.4\% accuracy for freshness classification, and 0.857 F1 score for apple detection. The framework provides an accessible and scalable alternative to multispectral UAV solutions, supporting practical precision agriculture on affordable hardware. 
\end{abstract}

\noindent\textbf{Keywords:}
Precision Agriculture; UAV; Deep Learning; Apple Disease; YOLOv8; ResNet-50; VGG-16.

\section{Introduction}
Apple orchards suffer from pest infestations, leaf diseases, and variable fruit quality, resulting in economic losses. Manual inspection is labour-intensive and prone to errors. UAV-based monitoring systems have emerged to solve this problem but they are often focused on single-task pipelines, such as only disease detection or only counting of fruits. 
This work addresses the lack of a unified or integrated UAV system that combines multiple tasks relevant to orchards. 
Existing studies have utilized YOLO variants for fruit detection, CNNs for leaf disease classifications, and UAV imaging with multispectral sensors, but in isolation, which limits scalability. However, no low-cost system integrates these things into one deployable pipeline.
The key contributions of this paper are as follows:
\begin{itemize}
    \item A unified UAV-based pipeline combining three relevant orchard monitoring tasks: leaf disease detection, apple freshness classification, and fruit localization
\item A cost-effective hardware architecture that simply uses ESP32-CAM imaging with Raspberry Pi edge inference.
\item A comparative evaluation of all three DL models across distinct orchard-relevant tasks using efficient pipelining methodology.
\item Real-time visualization using a custom web dashboard.

\end{itemize} 
Disease misdiagnosis, delayed detection, and inconsistent fruit quality assessment contribute to significant yield loss and market value reduction, making orchard monitoring a major economic challenge. A low-cost UAV-based solution can meaningfully reduce these losses for resource-limited growers.

\section{Literature Review}
Early works have used ML methods such as SVMs and KNNs \cite{8}, but CNN-based architectures have stood out in recent years due to their ability to extract hierarchical features from leaf imagery. Studies using VGG, GoogLeNet, and EfficientNet have demonstrated high efficiency in disease detection and rust across multiple crops \cite{1,2}. Residual networks like ResNet-50 have shown decent generalization for visually similar disease classes for leaf pathology tasks \cite{3}. Fruit quality assessment using RGB images has also gained traction, where CNN-based architectures have been applied to classify rotten vs fresh fruits across various fruits, leveraging shape, colour, and surface texture cues. VGG and baseline CNN-based work demonstrate strong performance in this regard \cite{4,5}.
UAV-mounted cameras have also been used for orchard monitoring, fruit counting, and yield estimation. Yield estimation has been performed using various YOLO variants, achieving excellent results even under challenging conditions such as occlusion, natural lighting, and dense clusters \cite{6,7}. While multispectral and hyperspectral sensors can capture physiological stress with higher precision but they remain expensive and require calibration, limiting adoption in small-scale systems. In contrast, recent studies show that RGB-based DL systems can achieve competitive performance at a fraction of the cost, making them suitable for real-world deployment. \cite{8,9}.

The existing literature addresses these issues in isolation, often leaving one or more problems unaddressed. No prior work has been done to integrate all three aspects into a unified, low-cost system capable of on-device inference. This work looks to bridge these gaps by combining three DL models into a cohesive end-to-end solution for orchard management.
\begin{figure}[ht]
    \centering
    \includegraphics[width=.5\linewidth]{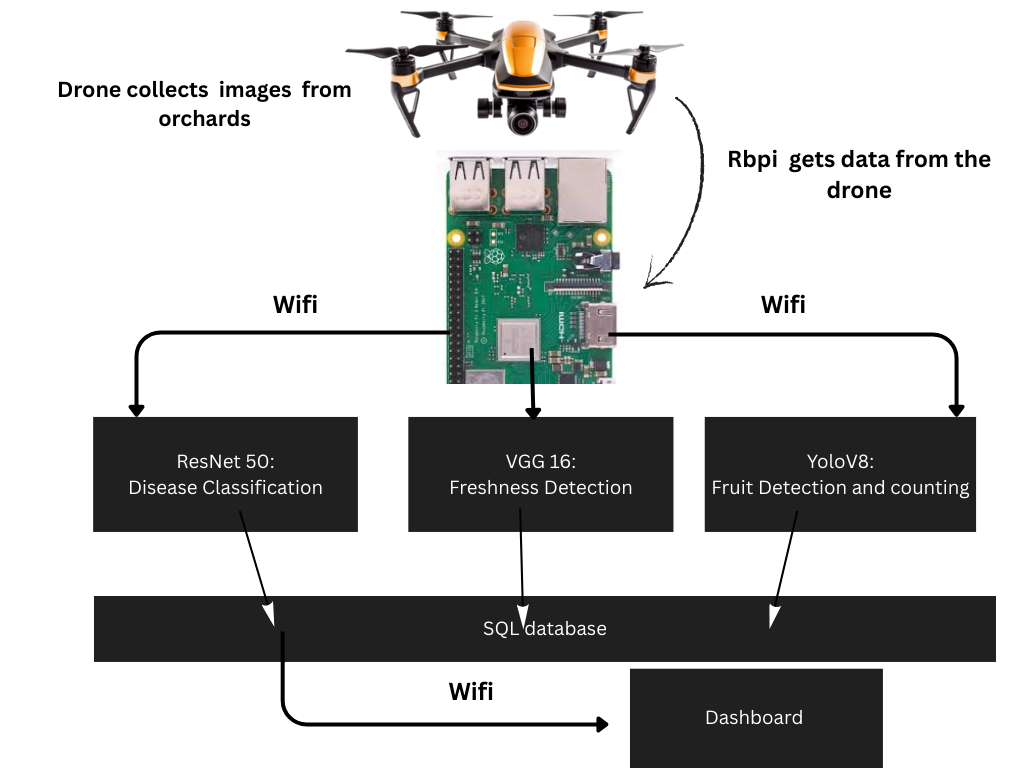}
    \caption{ Proposed pipeline }
\end{figure}
\\
\newline\noindent The proposed pipeline (Fig. 1) consists of :
\noindent  \subparagraph{\textbf{1.UAV-based image acquisition:}}
ESP32-CAM collects leaf and fruit images.
 \noindent \subparagraph{\textbf{2.Edge processing:}}
	The Raspberry Pi receives images via Wi-Fi, runs three DL models to process them, and stores them in a SQL database.
\noindent \subparagraph{\textbf{3.Three vision tasks:}}
\begin{itemize}
    \item ResNet-50: Apple disease classification
    \item VGG-16: Apple freshness classification
    \item YOLOv8: Fruit detection and localization
\end{itemize}
\section{Methodology} 
The study utilized three different image collections, each having approximately 10k samples: one for leaf disease classification, a second for apple freshness classification, and a third for apple detection. The datasets were partitioned using fixed stratified splits(train/val/test) of 80/20 for leaf disease, 64/16/20 for freshness, and 79/21 for the apple-detection dataset. The deep learning models were trained on these datasets. ResNet-50 was trained for 150 epochs using Adam optimizers and categorical cross-entropy loss on 224x224 inputs; VGG-16 for freshness classification, trained for 150 epochs on 256x256 inputs with data augmentation and Adam optimizer, and finally, YOLOv8 for detection and localization was trained for 300 epochs on 640x640 inputs.

The system implements a modular pipeline in which each model processes only data that is relevant to its use case. The Raspberry Pi performs the routing using simple heuristics based on flight context and camera framing (eg, high altitude framing indicates an orchard scene, hence YOLOv8 is used). This staged flow avoids redundant computation and avoids running all networks on every input, and enables the pipeline to execute sequentially on edge hardware. By decoupling tasks and structuring them into a lightweight processing chain, the architecture maintains low computational overhead while still maintaining multi-task orchard intelligence in real time.
\newpage
\section{Results}
The results of the experiments conducted on the aforementioned models are as follows 
\begin{table}[ht]
    \centering
    \begin{tabular}{|c|c|c|c|}
        \hline
        \textbf{Metrics} & \textbf{ResNet-50} & \textbf{VGG-16} & \textbf{YOLOv8} \\ \hline
        Validation Accuracy & 98.83 & 97.03 & - \\ \hline
        Validation Precision & 99 & 97 & 86.1 \\ \hline
        Validation Recall & 99 & 97 & 85.3 \\ \hline
        Validation F1-score & 99 & 97 & 85.7 \\ \hline
        mAP50 & - & - & 91 \\ \hline
        mAP50-95 & - & - & 63.1 \\ \hline
        Test Accuracy & 98.96 & 97.49 & - \\ \hline
        Test Precision & 99 & 98 & - \\ \hline
        Test Recall & 99 & 97 & - \\ \hline
        Test F1-score & 99 & 97 & - \\ \hline
    \end{tabular}
    \caption{ Models Performance Metrics}
\end{table}
\begin{figure}[ht]
    \centering
    \includegraphics[width=.7\linewidth]{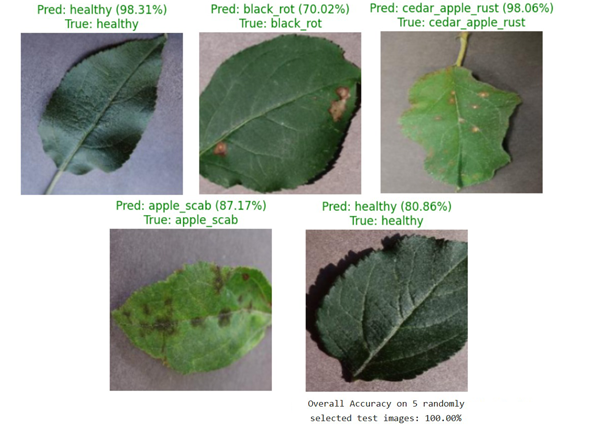}
    \caption{ Leaf-disease predictions from ResNet-50, correctly classifying healthy, black rot, cedar apple rust, and apple scab samples.}
\end{figure}
    \begin{figure}[ht]
 \centering
 \begin{minipage}[b]{0.28\textwidth}
  \includegraphics[width=.8\textwidth]{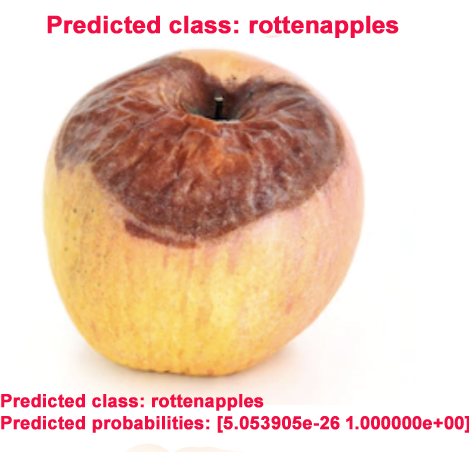}
 \end{minipage}%
 \begin{minipage}[b]{0.28\textwidth}
  \includegraphics[width=.8\textwidth]{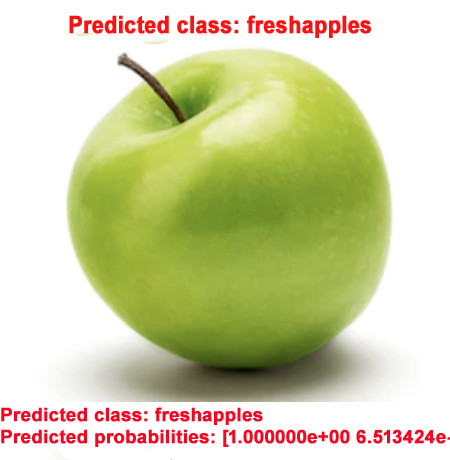}
 \end{minipage}
 \caption{ Freshness classification outputs from VGG-16, correctly identifying a rotten apple (left) and a fresh apple (right) with high confidence
}
\end{figure}

    \begin{figure}[ht]
 \centering
 \begin{minipage}[b]{0.27\textwidth}
  \includegraphics[width=.8\textwidth]{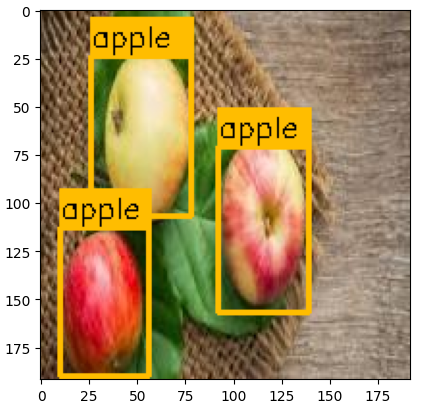}
 \end{minipage}%
 \begin{minipage}[b]{0.27\textwidth}
  \includegraphics[width=.8\textwidth]{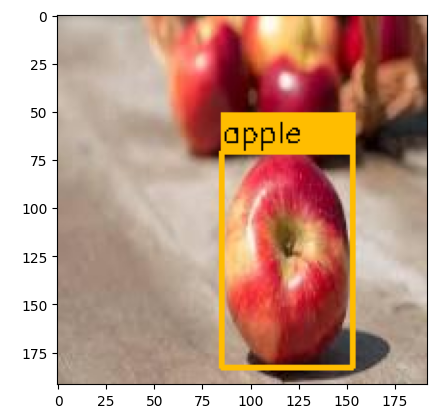}
 \end{minipage}
 \caption{Apple detection results from YOLOv8, accurately localizing apples under varying lighting and occlusion conditions.
}
\end{figure}
\newpage
\section{Discussion:}
The three model framework demonstrates that each architecture aligns with the specific visual complexity of its own task. ResNet-50, with its deep residual connections, is best suited for leaf disease detection because diseases manifest through subtle texture variations and distortions. With its ability to learn such finer hierarchical features, it allows for stronger generalization across multiple diseases, validating the use of such networks for pathological recognition.

VGG-16 worked for freshness classification due to its uniform convolutional hierarchy, which captures broad colour texture cues that are central to distinguishing a rotten apple from a fresh one. This indicates that model depth should match task complexity; deeper models excel at structural disease features, whereas moderate-depth CNNs are sufficient for appearance-based tasks. VGG-16 strikes an ideal balance between complexity and accuracy for binary color-texture classification, making it computationally efficient for edge devices. Together, these results highlight the importance of selecting architectures that are appropriately aligned with task-specific feature demands in agricultural vision.

YOLOv8's robust detection capabilities affirm the superiority of modern one-stage detectors in dynamic orchard environments. These environments present significant visual challenges, including heavy occlusion, rapid illumination shifts, and a wide range of fruit scales. The architecture's anchor-free detection head is pivotal, offering enhanced adaptability to the natural variance in apple sizes. Furthermore, its deep feature pyramid effectively bolsters detection robustness, particularly under partial occlusion. While performance is highly reliable across most scenarios, the primary areas for future optimization involve mitigating errors under the most extreme conditions, such as severe apple overlap or pronounced specular reflections from direct sunlight.
\section{Comparative analysis with existing literature}

Previous leaf disease classification employed CNN-based and transfer learning based approaches, with accuracies in the range of mid-to-high 90\% depending on dataset complexity and class similarity \cite{1}. Residual architectures are favored due to their ability to learn complex hierarchical features without degradation \cite{3}. The ResNet-50 model used in this work achieves results that are consistent with the findings, indicating that residual learning remains a feasible choice for fine-grained tasks such as disease classification.

Fruit freshness classification has also been explored previously with VGG-based and baseline CNNs, with results ranging from 92\%-96\% under controlled RGB imaging conditions \cite{4,5}. The VGG-16 model employed here achieves 97.4\% test accuracy, which aligns with the performance reported in existing literature. This shows that moderate depth CNNs provide a suitable balance between accuracy and computational efficiency, where freshness assessment relies on colour and surface texture cues.

Recent studies using YOLOv5 and YOLOv7 for orchard-scale fruit detection typically report mAP@50 scores in the 85–90\% range under natural lighting and partial occlusion \cite{6,7}, while YOLOv8 has shown improvements in the 90–93\% range on similar datasets\cite{12}. Our YOLOv8 model achieved 91\% mAP@50, placing it within the upper performance band reported in literature and confirming the suitability of next-generation one-stage detectors for orchard environments.

This work illustrates that task-aligned model integration, rather than architectural novelty alone, can yield robust and practically deployable orchard-monitoring systems.

\section{Conclusion}
 
This work presents a novel, unified UAV-based orchard monitoring framework that brings together disease detection, fruit freshness assessment, and apple localization within a single low-cost RGB pipeline. Unlike existing approaches that treat these tasks in isolation from each other or rely on expensive multispectral sensors, our system demonstrates that meaningful orchard management can be done through carefully selected DL models deployed entirely on accessible hardware. The solution can be extended to other fruits as well, such as oranges and bananas. Thus, this solution offers a practical and scalable alternative for real-world deployment without significant technological or economic overhead.

\section{Future scope}
To make the system even more affordable and efficient, future work will explore model optimization, reduced image transmission, and on-device inference. Temporal and spatial analysis and smarter flight-planning strategies can further streamline data collection and improve real-world deployment.
\newline \indent Incorporation of temporal consistency across video streams and more advanced routing strategies could further enhance robustness and computational efficiency by exploiting spatio-temporal redundancy in orchard-scale UAV data.
\newline \indent Moreover, future investigations might benefit from incorporating datasets that take into consideration real orchard-specific conditions, such as extreme illumination, dense canopy conditions, and effects due to seasonal changes, to further bolster system robustness. Additionally, Future work may include large-scale real-orchard flight validation and comparisons with multispectral sensing.
\newline \indent The present study deliberately focuses on RGB-based, edge-deployable pipelines to assess their feasibility under cost, computational, and accessibility constraints, thereby establishing a strong baseline for integrated orchard intelligence.

\bigskip 
\bigskip 
\bigskip 
\makebox[\linewidth]{\rule{5cm}{1pt}}
\bibliographystyle{unsrt}
\bibliography{REFNEW}
\end{document}